\documentclass{article}

\usepackage{times}
\usepackage[UKenglish]{babel}
\usepackage{varioref,amssymb,textcomp,bm,amsfonts}
\usepackage{amsmath}
\usepackage[pdftex]{graphicx}
\usepackage{tabularx}
\usepackage{natbib}
\usepackage{algorithm}
\usepackage{algorithmic}
\usepackage{hyperref}

\usepackage[accepted]{icml2016}

\graphicspath{{figures/}}

\begin{document}

\icmltitlerunning{Partial Reinitialisation for Optimisers}

\twocolumn[
  \icmltitle{Partial Reinitialisation for Optimisers}
  \icmlauthor{Ilia Zintchenko}{zintchenko@itp.phys.ethz.ch}
  \icmladdress{Theoretical Physics and Station Q Zurich, ETH Zurich,\\ 8093 Zurich, Switzerland}
  \icmlauthor{Matthew Hastings}{mahastin@microsoft.com}
  \icmladdress{Station Q, Microsoft Research,\\ Santa Barbara, CA
    93106-6105, USA}
  \icmladdress{Quantum Architectures and Computation Group, Microsoft
    Research,\\ Redmond, WA 98052, USA}
  \icmlauthor{Nathan Wiebe}{nawiebe@microsoft.com}
  \icmladdress{Quantum Architectures and Computation Group, Microsoft
    Research,\\ Redmond, WA 98052, USA}
  \icmlauthor{Ethan Brown}{ebrown@itp.phys.ethz.ch}
  \icmladdress{Theoretical Physics and Station Q Zurich, ETH Zurich,\\ 8093 Zurich, Switzerland}
  \icmlauthor{Matthias Troyer}{troyer@phys.ethz.ch}
  \icmladdress{Theoretical Physics and Station Q Zurich, ETH Zurich,\\ 8093 Zurich, Switzerland}

  \icmlkeywords{optimisation,clustering,neural networks,hidden markov}

  \vskip 0.3in  
]

\begin{abstract}
  Heuristic optimisers which search for an optimal configuration of
  variables relative to an objective function often get stuck in local
  optima where the algorithm is unable to find further improvement.
  The standard approach to circumvent this problem involves
  periodically restarting the algorithm from random initial
  configurations when no further improvement can be found. We propose
  a method of partial reinitialization, whereby, in an attempt to find
  a better solution, only sub-sets of variables are re-initialised
  rather than the whole configuration. Much of the information gained
  from previous runs is hence retained. This leads to significant
  improvements in the quality of the solution found in a given time
  for a variety of optimisation problems in machine learning.
\end{abstract}

\section{Introduction}

Multivariate optimisation is of central importance in industry and is
used for a range of problems from finding solutions which satisfy a
set of constraints to training classifiers in machine
learning. Despite the wide range of applications, nearly all of them
involve finding an optimal assignment of a set of variables with
respect to a cost function. An exact global optimum is, however,
rarely required and is in general significantly more costly to find
than a very good local optimum. Heuristic optimisers are therefore
ubiquitous for many industrial application
problems~\cite{zhang2001efficient,likas2003global,burges2011learning}. 

A heuristic optimiser generally starts from a random configuration of
variables and optimises this configuration until a local optimum is
found. For a convex optimisation problem this optimum is also the
global optimum and no further effort is required. However, problems
which are considered NP-hard have in general exponentially many such
local optima and hence the probability that a particular local optimum
is also the global one is exponentially suppressed. Nevertheless, such
problems often contain some amount of structure which makes their
solution useful. This structure can be used to obtain configurations
of a given cost much faster than random guessing. As such, each
subsequent query to the cost function makes use of preceding queries
to make a more informed guess of an optimal configuration.

A strategy that is frequently used to escape a local optimum is to
restart the optimisation process from a new random initial
configuration. Repeating this process multiple times, the local
optimum with the lowest cost is then returned, which is with high
probability better and with certainty no worse than the initial
configuration. Although such restarts allow the optimiser to get out
of local optima, different restarts are also completely decoupled from
each other. That is, information which was learned about the structure
of the problem in one restart is not passed on to the next and has to
be relearned from scratch. Hence, this way of running an optimiser is
effectively coarse grained random guessing, but where each guess is
further improved towards a local optimum.

Here we introduce a general approach to address this problem which
allows an optimiser to find the global optimum with high probability
in a single run.

\section{Algorithm}

\begin{figure}[t!]
  \centering
  \includegraphics[width=1.0\columnwidth]{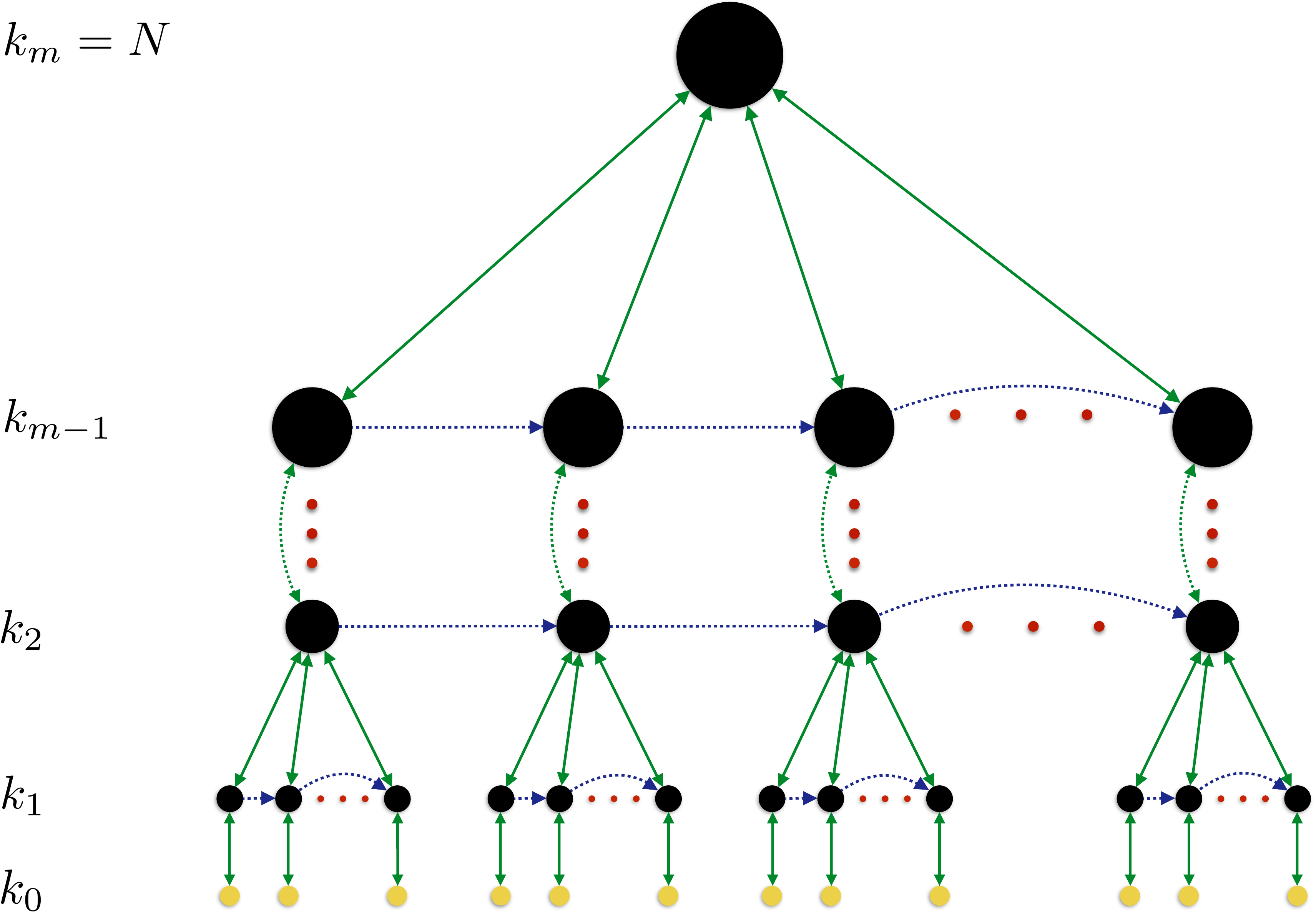}
  \caption{An illustration of the partial reinitialisation
    algorithm. On each level $l$, sub-sets of $k_l$ variables are
    reinitialised (\emph{black circles}) such that $(k_m = N) >
    k_{m-1} > \ldots > k_0$. On the bottom level the $k_0$-optimal
    optimiser is called (\emph{yellow circles}). Each level starts
    with the most optimal configuration of its parent and
    reinitialises a sub-set of $k_l$ variables before calling the
    optimiser on the next level (\emph{green arrows}). Checkpoints of
    the most optimal configuration found are kept on each level with a
    flow from left to right (\emph{blue arrows}).}
  \label{fig:hierarchy}
\end{figure}

Lets assume we have a heuristic which picks sub-sets of variables.
Also, let us define $k_0$-optimality of an optimiser such that for any
configuration it returns, reinitialising the variables in a typical
sub-set smaller than $k_0$ found by this heuristic does not get the
configuration out of the local optimum. That is, the optimiser would
just return the same configuration again. However, reinitialising
sub-sets of $k_1 > k_0$ variables may allow the optimiser to find a
set of new local optima, some of which may be worse or better than the
current one. Starting from $k_1 = k_0$, lets increase $k_1$ until a
better local optimum is reachable for a typical sub-set picked by our
heuristic. If the current local optimum is good, that is, the number
of better local optima is negligible, increasing $k_1$ further would
only reduce the chance of finding a better local optimum. Hence,
except in the very beginning of the optimisation process, the
optimiser has a higher chance of finding a better local optimum after
re-initialising sub-sets of only $k_1$ rather than $N$ variables,
where $k_0 < k_1 \ll N$.

As sub-sets of $k_1$ variables are reinitialised and the optimiser
called after each re-initialisation, the configuration becomes
$k_1$-optimal with high probability and the chance of finding a better
local optimum decreases. To prevent the optimiser from getting stuck
in the $k_1$-optimum, sub-sets of $k_2 > k_1$ variables can be
re-initialised. In turn, to get out of $k_2$-local optima, sub-set of
$k_3 > k_2$ variables can be re-initialised and so on. Repeating this
process iteratively, each time increasing the size of the sub-set
until $(k_m=N)$, the configuration becomes $N$-optimal, which is the
global optimum with high probability. This process can hence refine a
local optimizer into a global optimizer. We call this approach partial
re-initialisation with the following algorithm (see also
Fig.~\ref{fig:hierarchy}).

\begin{algorithm}[H]
  \caption{Partial reinitialisation}
  \label{alg:pinit}
  \begin{algorithmic}

    \STATE {\bfseries Input:} current level $l$, number of
    reinitialisations $M_l$ and number of variables for each
    reinitialisation $k_l$

    \IF{$l = 0$}
    \STATE call heuristic optimiser on $\mathbf{x}$
    \ELSE

    \STATE $\mathbf{x}_0 \gets \mathbf{x}$
    \STATE reinitialise sub-set of $k_l$ variables in $\mathbf{x}$
    \FOR{$i \in \{1\ldots M_l\}$}
    \STATE call partial reinitialisation on level $l-1$
    \ENDFOR

    \IF{cost($\mathbf{x}$) $>$ cost($\mathbf{x_0}$)}
    \STATE $\mathbf{x} \gets \mathbf{x_0}$
    \ENDIF

    \ENDIF
  \end{algorithmic}
\end{algorithm}
With $m$ levels in the hierarchy, the algorithm is started from the
$m^\textnormal{th}$ level. The configuration is denoted by
$\mathbf{x}$ and the checkpoints by $\mathbf{x_0}$; the checkpoints
are optimal configurations found thus far, to which the algorithm
reverts if no improved configuration is found by partial
reinitialization. At each level $l$, $M_l$ reinitialisations of $k_l$
variables are performed and $(k_m = N) > k_{m-1} > \ldots > k_0$. The
complexity of a single run of the algorithm is hence
$\mathcal{O}(\prod_l M_l)$.

Not only the number of variables, but also the heuristic which picks
the variables in a sub-set is important. The simplest approach is to
pick variables at random. However, if variables are chosen according
to a problem-specific heuristic, the probability that re-initialising
a sub-set of a given size leads to a more optimal configuration can be
maximised. One approach is to pick sub-sets such that the optimality
of variables within the set depends on the values of the other
variables in the set as much as possible. This could increase the
chance of getting out of a local optimum by reducing the number of
constraints on the sub-set from the rest of the system.

If the outcome of the heuristic optimiser does not directly depend on
the initial configuration, but also on a random seed, it could be used
to optimise only the variables within a sub-set, while the other
variables in the problem are kept fixed. Such an approach was employed
in~\cite{groups} for finding ground states of Ising spin glasses with
simulated annealing and showed significant speedup relative to
conventional global restarts.

If the optimisation problem is over the space of continuous variables,
the concept of partial reinitialisation can be extended to partially
reinitialising each variable in addition to sub-sets of
variables. That is, rather than setting a variable to a random value
within a pre-defined domain, it can be perturbed by, for example,
adding noise with some standard deviation $\sigma$, that is
\begin{equation}
x = \alpha x +(1-\alpha)\mathcal{N}(\mu,\sigma)
\end{equation}
where $\mu$ is the mean and $\sigma$ is the standard deviation of the
distribition $\mathcal{N}$. Hence, we can either fully reinitialise
sub-sets of variables, add small perturbations to all variables, or
combine the two and partially perturb sub-sets of variables, which
could further improve performance. For simplicity we use $\alpha = 0$
in the benchmarks below.

\subsection{Choosing $M_\ell$}

Although $M_\ell$ can be chosen empirically to optimise the
performance of an optimiser over a set of randomly chosen optimization
problems, there is also a theoretical basis behind the
choice. $M_\ell$ should be chosen to ensure that the probability of
being in a local optima (with respect to reinitiasation of $k_\ell$
variables from the input) is maximised. We call such a configuration
\emph{probabilistically $k_{\ell}$--optimal}.

If we require that the probability that a reinitialization of $k_\ell$
variables does not improve the optimum is less than $1-\delta$ and
assert that the probability that a given reinitialisation improves the
objective function is $P \ge \epsilon$ then it suffices to
take~\cite{donmez2009local}
\begin{equation}
  M_\ell \ge \lceil \ln(\delta)/\ln(1-\epsilon) \rceil.
\end{equation}
Thus the values of $\delta$ and $\epsilon$ specify the value of
$M_\ell$ needed to conclude that the local optimum is
probabilistically $k_\ell$-optimal within a margin of
error. Furthermore, if we consider $1-\epsilon$ to be a constant then
only a logarithmic number of samples are needed to conclude with high
probability that the value is $k_\ell$-optimal. If $\epsilon$ is small
on the other hand, such as in unstructured search, then the number of
requisite reinitialisations can be exponential in the number of
variables. Thus the value of $\epsilon$ taken makes an implicit
assumption about the structure and complexity of the problem.

\section{Benchmarks}

On a set of machine learning problems we study the advantage of
partial reinitialisation compared to the standard full
reinitialisation for finding optimum model parameters. These problems
are picked from the ones with which we have most experience. The size
of each problem was chosen to be large enough such that finding the
optimal configuration is non-trivial using the respective standard
algorithms.

It is worth noting that in machine learning, the local optimum is
sometimes sufficient or even more desired than a global optimum due to
overtraining~\cite{LeCun2015}. However, as we show below, partial
reinitialisation improves not only the quality of the final solution,
but also the speed at which a local optium of a given quality is
obtained. Also, although in general more expensive, overtraining can
be reduced by using the classicifation accuracy on random sub-sets of
the training data as a cost function.

For simplicity we use only one level in the hierarchy between a full
reinitialisation and calling the heuristic optimiser. That is, for
each full reinitialisation, multiple reinitialisations of sub-sets of
variables are perfomed. To maintain generality, we choose sub-sets at
random for all benchmark problems.

The parameters in the benchmarks, such as the size of each subset
(denoted by $k_1$) and the number of partial reinitialisations
(denoted by $M_1$) which are done within each full reinitialisation,
were manually optimised and are not the true optima for the respective
performance metrics.

As a performance measure for each benchmark we use the most optimal
cost obtained after a given elapsed time or number of iterations
averaged over multiple runs with different random initial
states. Elapsed time was measured on Intel Xeon E5-2660 v2 processors.

\subsection{Training hidden Markov models}

\begin{figure}[!h]
  \centering
  \includegraphics[width=1.0\columnwidth]{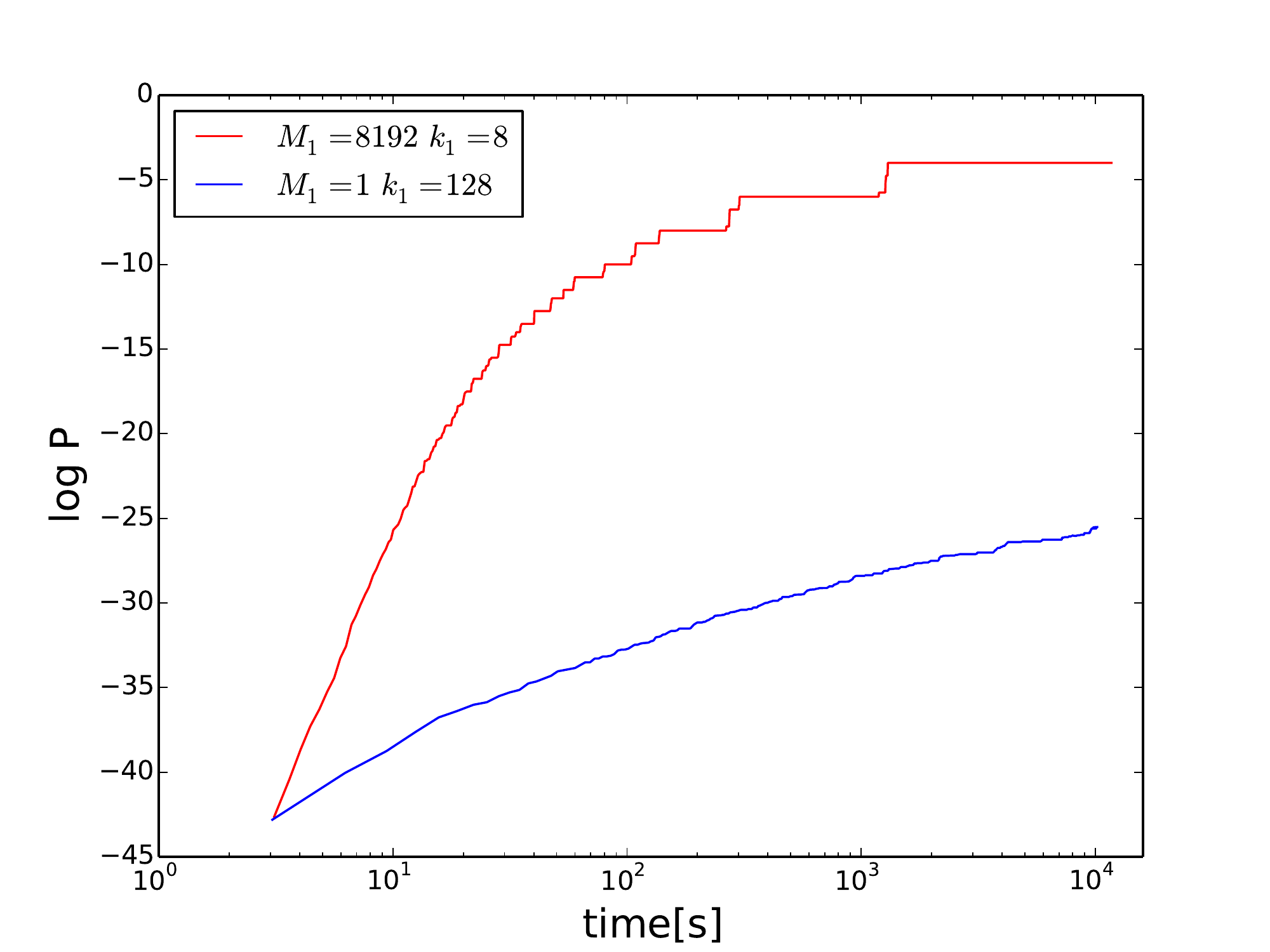}
  \caption{Learning random $128$-bit sequences using the Baum-Welch
    algorithm. Median log-likelihood to observe the sequence within
    the model vs. elapsed training time over $1000$ runs with
    different random seeds is shown. Full reinitialisations are
    \emph{blue} and partial reinitialisations \emph{red}.}
  \label{fig:hmm}
\end{figure}

Learning temporal patterns in a signal is of central importance in a
wide range of fields including speech recognition, finance and
bioinformatics. A classic method to model such systems is hidden
Markov models (HMM), which are based on the assumption that the signal
follows a Markov process. That is, the future state of the system
depends solely on the present state without any memory of the
past. This assumption turns out to be surprisingly accurate for many
applications.

In discrete HMMs, which we will consider here, the system can be in
one of $N$ possible states hidden from the observer. Starting from a
discrete probability distribution over these states, as time evolves
the system can transition between states according to an $N\times N$
probability matrix $A$. Each hidden state can emmit one of $M$
possible visible states. The model is hence composed of three parts:
the initial probability distribution of length $N$ over the hidden
states; the $N\times N$ transition matrix between hidden states; the
$N\times M$ emission matrix from each hidden state into $M$ possible
visible states. During training on a given input sequence, these
matrices are optimised such as to maximise the likelihood for this
sequence to be observed.

The standard algorithm for training HMMs is the Baum-Welch
algorithm~\cite{Rabiner89atutorial}. It is based on the the
forward-backward procedure, which computes the posterior marginal
distributions using a dynamic programming approach. The model is
commonly initialised with random values and optimised to maximise the
expectation of the input sequence until convergece to a local
optimum. To improve accuracy, multiple restarts are usually
performed. Over the sequence of restarts, we will investigate if
partial reinitialisation can improve the convergence rate towards a
global optimum.

As a benchmark we choose learning a random $128$-bit string. Although
artificial, this problem is very simple and has few free
parameters. At the same time, finding a model which accurately
represents the sequence is non-trivial starting from random transition
and emission matrices. If the number of hidden states is at least as
large as the length of the bit-string, the model can be trained until
that bit-string is observed with certainty within the model. Here we
will use two visible states and the same number of hidden states as
bits in the input. For generality we do not impose any restrictions on
the transition matrix.

Each full reinitialisation starts with random emission and transition
matrices. In a partial reinitialisation only the elements in the
matrices corresponding to a sub-set of hidden states are
reinitialised. We found about $8000$ reinitialisations of $8$
variables within each global restart to be optimal and leads to a much
more rapid increase in accuracy with training time than with full
reinitialisations (see Fig.~\ref{fig:hmm}).

\subsection{Clustering with k-means}

\begin{figure}[t!]
  \centering
  \includegraphics[width=1.0\columnwidth]{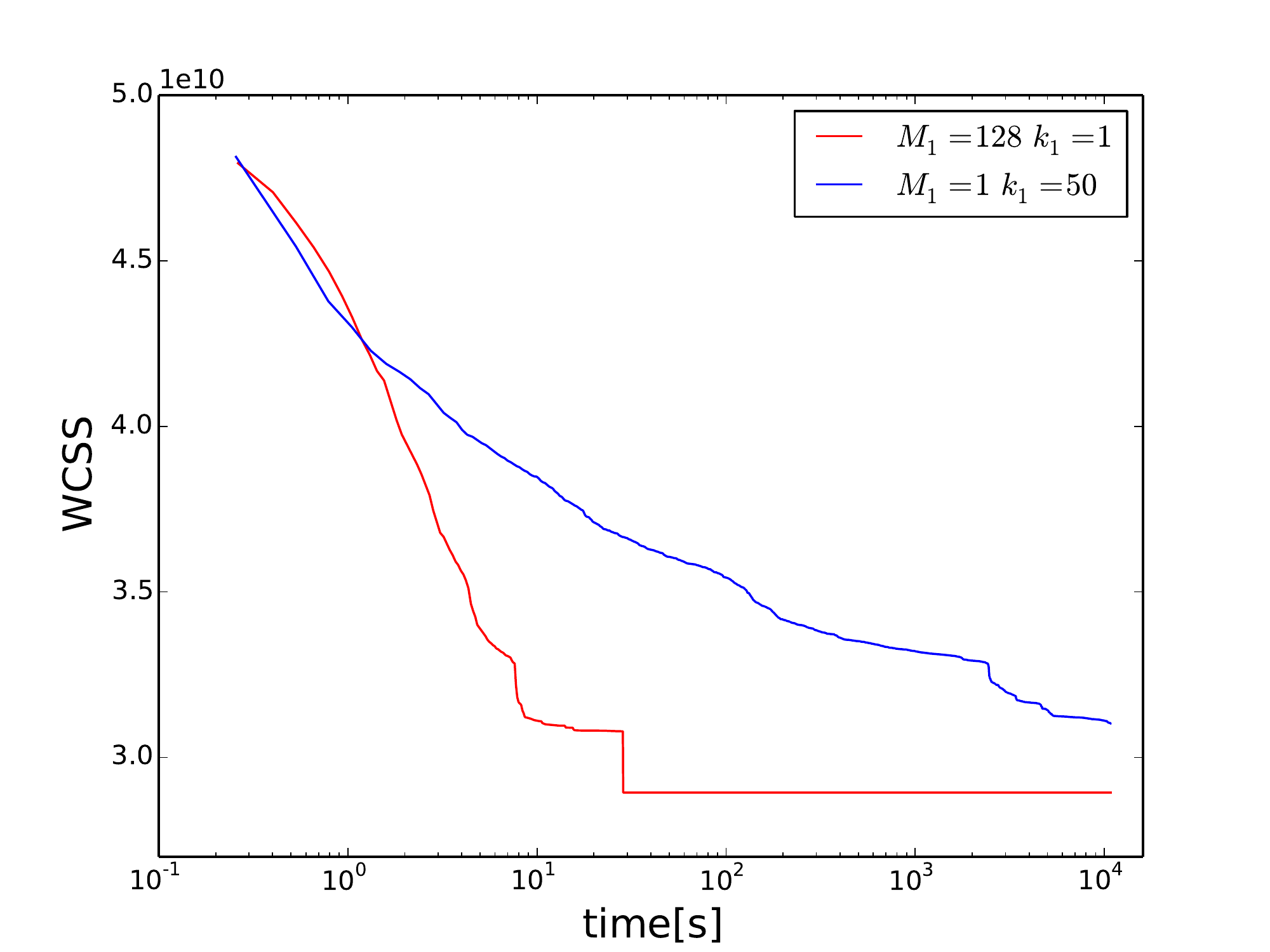}
  \caption{The benchmark is the A3 set from~ \cite{clustering}. Median
    within-cluster sum of squares (WCSS) and elapsed time over $1000$
    runs of the k-means algorithm is shown. The \emph{blue} curve is
    full reinitialisations and the \emph{red} curve is partial
    reinitialisations. }
  \label{fig:a3_benchmark}
\end{figure}

Dividing objects into clusters according to a similarity metric is of
central importance in data analysis and is employed ubiquitously in
machine learning. Given a set of points in a finite-dimensional space,
the idea is to assign points to clusters in such a way as to maximise
the similarities within a cluster and minimise the similarities
between clusters. Similarity can be defined in many ways, depending on
the particular needs of the application. Here we use the Euclidean
distance.

One of the most widely used algorithm for finding such clusters is the
k-means algorithm~\cite{macqueen1967}. K-means searches for an
assignment of points to clusters such as to minimise the
within-cluster sum of square distances to the center. That is, it
seeks to minimise the following cost function
\[
\sum_{i=1}^k \sum_{j\in C_i}\|\mathbf{x}_j - \mu_i\|^2
\]
Starting from a random initialisation of centers, each iteration
proceeds in two stages. First all points are assigned to the nearest
cluster center. In the second part each center is picked to be the
Euclidean center of its cluster. This is repeated until convergence to
a local optimum. As the cost is never increased, convergence is
guaranteed. It has been shown that there exists problems on which
k-means converges in exponentially many iterations in the number of
points~\cite{worst_kmeans}. The smoothed running time is, however,
polynomial~\cite{smoothed_kmeans} and it typically converges very
quickly. Similar to the Baum-Welch algorithm, multiple global
reinitialisations of centers are often performed to improve the
quality of the clusters.

Our benchmark problem is set A3 in a standard clustering data
set~\cite{clustering}. The clusters have points sampled from Gaussian
distributions. There are $7500$ points around $50$ uniformly placed
centres. Each full reinitialisation starts with an intial assignment
of all centers to random points using Forgy's
method~\cite{forgy65}. In a partial reinitialisation only a single
cluster center is reinitialised. We found about $100$ partial
reinitialisations for each global restart to be optimal.

Except in the very beginning, a given quality of clusters is obtained
significantly quicker with partial rather than full reinitialisation
(see Fig.~\ref{fig:a3_benchmark}). The reason full reinitialisation is
here more efficient at finding low quality clusters is because in the
regime where the cost is high, larger strides in reducing the cost can
be made by just random guessing than optimising a bad global restart
further with partial reinitialsiations. However, as the cost threshold
becomes harder to reach, partial reinitialisation quickly becomes more
efficient.

\subsection{Clustering with k-medoids}

\begin{figure}[!h]
  \centering
  \includegraphics[width=1.0\columnwidth]{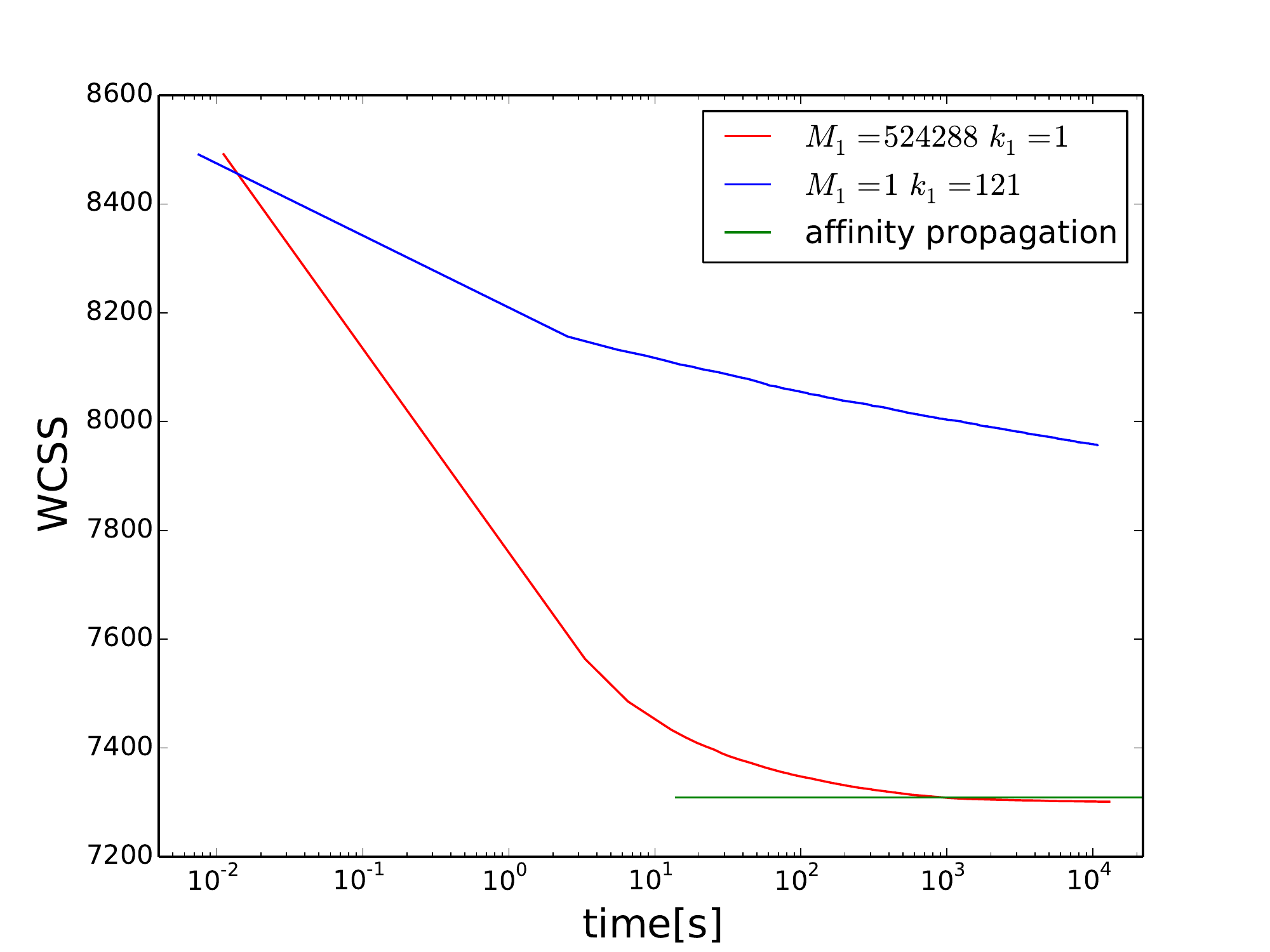}
  \caption{The benchmark problem is clustering $900$ images of faces
    from the Olivetti face database~\cite{face_database} into $121$
    clusters. Affinity propagation (\emph{green} curve) compared to
    k-medoids with full (\emph{blue} curve) and partial (\emph{red}
    curve) reinitialisation. The affinity propagation benchmark with
    coresponding timings was done using the web
    interface~\cite{affinity_web}.}
  \label{fig:faces_benchmark}
\end{figure}

An often more robust approach, called k-medoids, for clustering data
is to pick the best cluster center to be one of the points in the
cluster rather than the Euclidean center. The standard algorithm for
finding such clusters is partitioning around medoids
(PAM)~\cite{pam}. Similar to the k-means algorithm, PAM iterates
between assigning points to their closest center and finding the best
center of each cluster until convergence.

As the centers are constrainted to the positions of the points and no
cluster assignment can occur twice during a run, the trivial upper
bound on the number of iterations until convergence is $N\choose{k}$
for $N$ points and $k$ clusters. Similar to k-means, however, it
typically converges very quickly.

The benchmark problem we use here is clustering $900$ images of faces
from the Olivetti database~\cite{face_database}. This dataset has been
used previously to benchmark a novel method, called affinity
propagation~\cite{affinity_propagation}, for finding k-medoids
clusters against PAM. Here we repeat this benchmark, but also compare
to PAM with partial reinitialisations.

We chose to find $121$ clusters as this was roughly the regime where
affinity propagation had the largest advantage over
k-medoids~\cite{affinity_propagation}. Within each gloal restart, we
found that about $0.5\cdot 10^6$ reinitialisations of only a single
randomly chosen cluster center is optimal. We use a pre-computed
similarity matrix, available from~\cite{face_database_similarities},
the entries of which are the pixel-wise squared distances between the
images after some pre-processing.

Affinity propagation indeed quickly finds much better clusters than
PAM with full reinitialisations. However, after some time PAM with
partial reinitialisations finds even better clusters which keep
improving even further with time (see
Fig.~\ref{fig:faces_benchmark}). Full reinitialisation finds better
cluters in the beginning for similar reasons as in k-means.

\subsection{Training Bolzmann machines}

\begin{figure*}[t!]
  \centering
  \includegraphics[width=0.49\linewidth]{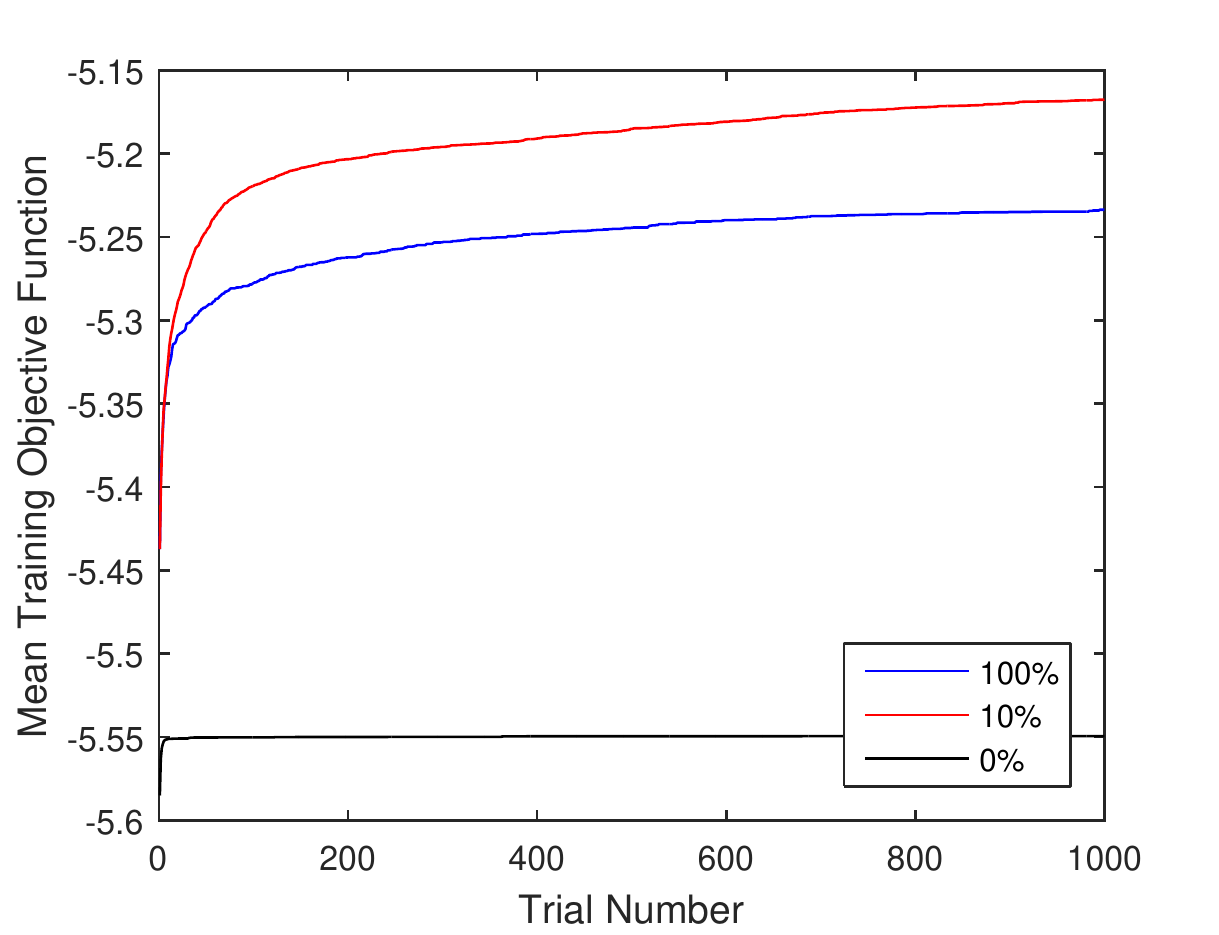}
  \includegraphics[width=0.49\linewidth]{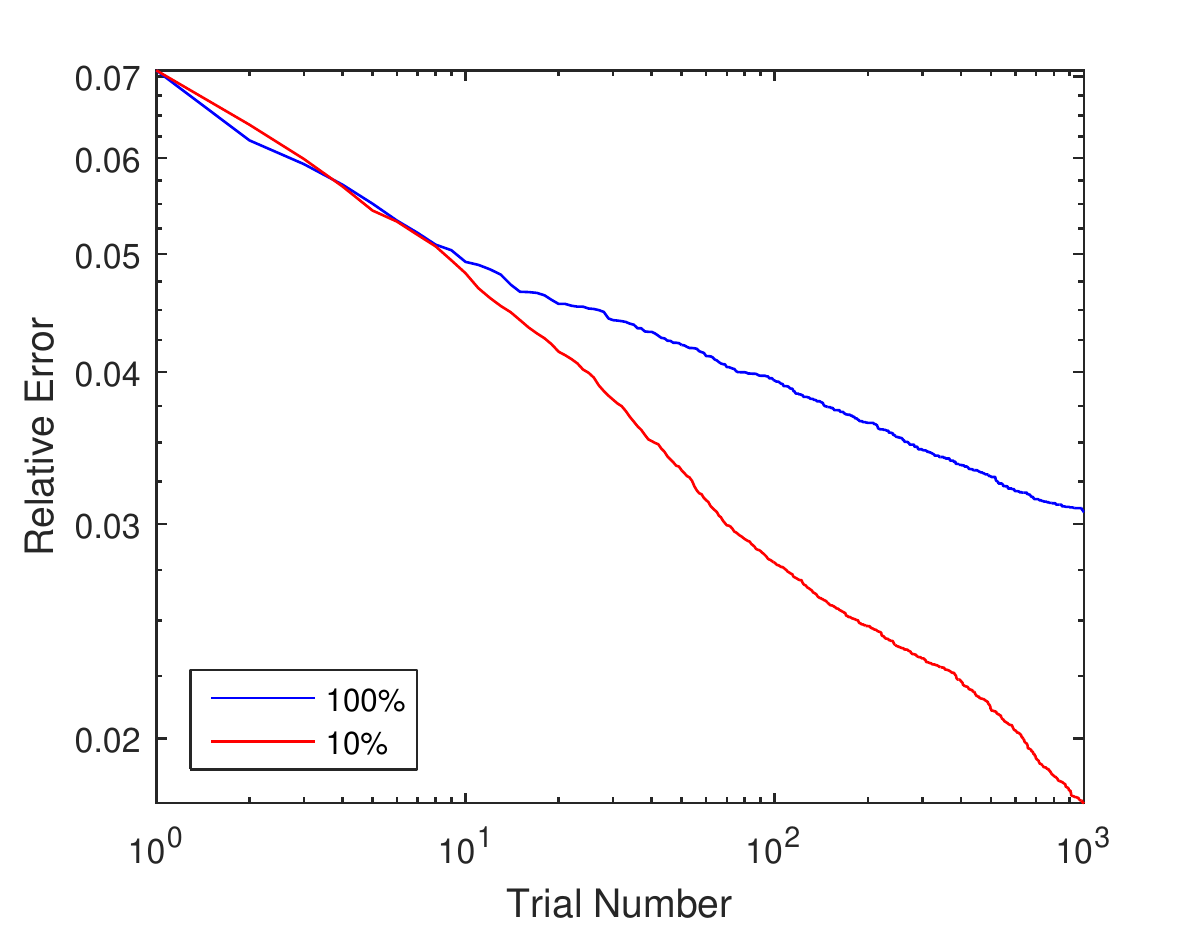}
  \caption{Training of Restricted Bolzmann Machine (RBM) with
    $\lambda=0$ (left) absolute values of objective functions and
    (right) relative difference between mean $O_{\rm ML}$ and the
    ``approximate global optima'' found by maximizing $O_{\rm ML}$
    over all experimental runs.}
  \label{fig:bm}
\end{figure*}

Boltzmann machines are a class of highly generalizable models, related
to feed-forward neural networks, that are have proven to be very
useful for modeling data sets in many areas including speech and
vision~\cite{bengio2009learning,hinton2002training,salakhutdinov2009deep}.
The goal in Boltzmann machine training is not to replicate the
probability distribution of some set of training data, but rather to
identify patterns in the data set and generalize them to cases that
have not yet been observed.

The Boltzmann machine takes the following form: Let us define two
layers of units, which we call the visible layer and the hidden
layer. The visible units comprise the input and output of the
Boltzmann machine and the hidden units are latent variables that are
marginalized over in order to generate the correlations present in the
data.  We call the vector of visible units $\mathbf{v}$ and the vector
of hidden units $\mathbf{h}$. These units are typically taken to be
binary and the joint probability of a configuration of visible and
hidden units is
\begin{equation}
  P(\mathbf{v},\mathbf{h})=\frac{\exp\left(-E(\mathbf{v},\mathbf{h})
    \right)}{Z},\label{eq:gibbs}
\end{equation}
where $Z$ is a normalization factor known as the partition function and 
\begin{equation}
  E(v,h) =- \mathbf{v}\cdot \mathbf{a} -\mathbf{h}\cdot \mathbf{b} -
  \mathbf{v}^T W \mathbf{h},
\end{equation}
is the energy. Here $W$ is a matrix of weights that models the
interaction between pairs of hidden and visible units. $a$ and $b$ are
vectors of biases for each of the units. This model can be viewed as
an Ising model on a complete bipartite graph that is in thermal
equilibrium.

This model is commonly known as a Restricted Boltzmann Machine
(RBM). Such RBMs can be stacked to form layered Boltzmann machines
which are sometimes called deep Boltzmann machines. For simplicity we
will focus on training RBMs since training deep Boltzmann machines
using popular methods, such as contrastive divergence training,
involves optimising the weights and biases for each layered RBM
independently.

The training process involves optimising the maximum likelihood
training objective, $O_{\rm ML}$, which is
\begin{equation}
  O_{\rm ML} = \mathbb{E}_{\mathbf{v}\in \mathbf{x}_{\rm train}}
  \Big(\ln[\mathbb{E}_{h} P(\mathbf{v},\mathbf{h})]\Big)-\frac{\lambda
    \sum_{ij} W_{ij}^2}{2},
\end{equation}
where $\lambda$ is a regularization term introduced to prevent
overfitting. The exact computation of the training objective function
is \#P hard, which means that its computation is expected to be
intractable for large RBMs under reasonable complexity theoretic
assumptions.

Although $O_{\rm ML}$ cannot be efficiently computed, its derivatives
can be efficiently estimated using a method known as contrastive
divergence. The algorithm, described in detail
in~\cite{hinton2002training}, uses a Markov chain algorithm that
estimates the expectation values of the hidden and visible units which
are needed to compute the derivatives of $O_{\rm ML}$. Specifically,
\begin{equation}
  \frac{\partial O_{\rm ML}}{\partial W_{ij}}= \langle v_i h_j
  \rangle_{\rm data}-\langle v_i h_j \rangle_{\rm model}-\lambda
  W_{ij}.
\end{equation}
Here $\langle \cdot \rangle_{\rm data}$ denotes an expectation value
over the Gibbs distribution of~\eqref{eq:gibbs} with the visible units
clamped to the training data and the $\langle \cdot \rangle_{\rm
  model}$ denotes the unconstrained expectation value. The derivative
with respect to the biases is similar. Locally optimal configurations
of the weights and biases can then be calculated by stochastic
gradient ascent using these approximate gradients.

As a benchmark we will examine small synthetic examples of Boltzmann
machines where the training objective function can be calculated
exactly. Although this differs from the task based estimates of the
performance of a Bolzmann machine, we focus on it because the
contrastive divergence training algorithm formally seeks to
approximate the gradients of this objective function. As such the
value of the training objective function is a more natural metric of
comparison for our purposes than classification accuracy for an
appropriate data set.

The training set consists of four distinct functions:
\begin{align}
  &[x_1]_j = \begin{cases}1 & j=1,\ldots,\lfloor n_v/2\rfloor\\ 0 &
    {\rm otherwise} \end{cases}\nonumber\\ &[x_2]_j = j \mod
  2\label{eq:4datavectors},
\end{align}
and their bitwise negations. In order to make the data set more
challenging to learn, we add $10\%$ Bernoulli noise to each of the
training examples. One hundred training examples were used in each
instance with a learning rate of $0.01$ and $10^5$ epochs per
reinitialization. We take the model to be an RBM consisting of $8$
visible units and $10$ hidden units. Finally, rather than
reinitialising a fixed number of weights at each iteration we update
each weight with a fixed probability. This formally differs from
previous approaches, but performs similarly to reinitialising a
constant fraction. Upon initialisation, each variable is drawn from a
zero-mean Gaussian distribution.

We see in Figure~\ref{fig:bm} that partial reinitialization
accelerates the process of estimating the global optima for
contrastive divergence training. We find that the optimal probability
of reinitializing a weight or bias in the BM is roughly $10\%$. This
tends to lead to substantial reductions in the number of
reinitializations needed to attain a particular objective function. In
particular, for $\lambda=0$ if $10\%$ of the variables are reset at
each stage then it only takes $58$ resets on average to obtain the
same training objective function as the strategy that resets every
variable attains with $1000$ resets.  Furthermore, we see evidence
that suggests a possible polynomial advantage of the $10\%$ strategy
over the traditional random resetting strategy.

Strong regularization is expected to lead to a much simpler landscape
for the optimisation since as $\lambda \rightarrow \infty$ the
optimisation problem becomes convex. We find that for $\lambda=0.01$
that the difference between the partial and total reinitisation
strategies becomes on the order of $0.01\%$.  Although partial
reinitialising $10\%$ of the variables continues to be the superior
method, these small differences could also arise solely from the
stochastic nature of contrastive divergence training and the fact that
it takes fewer epoches for the partial reinitialization to approach
the local optima.

It is evident from these benchmarks that partial reinitialization can
substantially reduce the complexity of finding an approximate global
optimum for Boltzmann machines. Consequently, we expect that this
approach will also work for optimising Boltzmann training for large
task--based problems such as MNIST digit classification.  Similarly,
these improvements should be able to be leveraged in other forms of
neural net training, such as feedforward convolutional neural nets.

\section{Conclusion}

We introduced a general purpose approach, which we termed partial
re-initialisation, to significantly improve the performance of an
optimiser. We numerically explored comparisons to state of the art
algorithms on a range of optimisation problems in machine learning and
although we only used the most basic version of our algorithm, with a
single additional level in the hierarchy and picking sub-sets of
variables at random, the advantage over the standard full
reinitialisation is substantial. We expect a hierarchy with multiple
levels and with sub-sets picked according to more advanced
problem-specific heuristics to lead to even further improvements.

\section*{Acknowledgments} 

We would like to thank A. Supernova for inspiration and J. Imriska,
H. Katzgraber, A. Kosenkov, A. Soluyanov, K. Svore, D. Wecker for
fruitful discussions. This paper is based upon work supported in part
by ODNI, IARPA via MIT Lincoln Laboratory Air Force Contract
No. FA8721-05-C- 0002. The views and conclusions contained herein are
those of the authors and should not be interpreted as necessarily
representing the official policies or endorsements, either expressed
or implied, of ODNI, IARPA, or the U.S. Government. The
U.S. Government is authorized to reproduce and distribute reprints for
Governmental purpose notwithstanding any copyright annotation thereon.

\bibliographystyle{icml2016}
\bibliography{paper}

\begin{thebibliography}{20}
\providecommand{\natexlab}[1]{#1}
\providecommand{\url}[1]{\texttt{#1}}
\expandafter\ifx\csname urlstyle\endcsname\relax
  \providecommand{\doi}[1]{doi: #1}\else
  \providecommand{\doi}{doi: \begingroup \urlstyle{rm}\Url}\fi

\bibitem[aff(2015)]{affinity_web}
Affinity propagation web app.
\newblock \url{http://www.psi.toronto.edu/affinitypropagation/webapp/}, 2015.
\newblock Accessed: 2015-08-18.

\bibitem[clu(2015)]{clustering}
Clustering datasets.
\newblock \url{http://cs.joensuu.fi/sipu/datasets/}, 2015.
\newblock Accessed: 2015-05-25.

\bibitem[fac(2015{\natexlab{a}})]{face_database}
Olivetti face database.
\newblock \url{http://www.psi.toronto.edu/affinitypropagation/Faces.JPG},
  2015{\natexlab{a}}.
\newblock Accessed: 2015-08-18.

\bibitem[fac(2015{\natexlab{b}})]{face_database_similarities}
Olivetti face database.
\newblock \url{http://www.psi.toronto.edu/index.php?q=affinity\%20propagation},
  2015{\natexlab{b}}.
\newblock Accessed: 2015-09-11.

\bibitem[Arthur et~al.(2009)Arthur, Manthey, and R\"{o}glin]{smoothed_kmeans}
Arthur, David, Manthey, Bodo, and R\"{o}glin, Heiko.
\newblock k-means has polynomial smoothed complexity.
\newblock In \emph{Proceedings of the 2009 50th Annual IEEE Symposium on
  Foundations of Computer Science}, FOCS '09, pp.\  405--414, Washington, DC,
  USA, 2009. IEEE Computer Society.
\newblock ISBN 978-0-7695-3850-1.
\newblock \doi{10.1109/FOCS.2009.14}.
\newblock URL \url{http://dx.doi.org/10.1109/FOCS.2009.14}.

\bibitem[Bengio(2009)]{bengio2009learning}
Bengio, Yoshua.
\newblock Learning deep architectures for ai.
\newblock \emph{Foundations and trends{\textregistered} in Machine Learning},
  2\penalty0 (1):\penalty0 1--127, 2009.

\bibitem[Burges et~al.(2011)Burges, Svore, Bennett, Pastusiak, and
  Wu]{burges2011learning}
Burges, Christopher~JC, Svore, Krysta~Marie, Bennett, Paul~N, Pastusiak,
  Andrzej, and Wu, Qiang.
\newblock Learning to rank using an ensemble of lambda-gradient models.
\newblock In \emph{Yahoo! Learning to Rank Challenge}, pp.\  25--35, 2011.

\bibitem[Donmez et~al.(2009)Donmez, Svore, and Burges]{donmez2009local}
Donmez, Pinar, Svore, Krysta~M, and Burges, Christopher~JC.
\newblock On the local optimality of lambdarank.
\newblock In \emph{Proceedings of the 32nd international ACM SIGIR conference
  on Research and development in information retrieval}, pp.\  460--467. ACM,
  2009.

\bibitem[Forgy(1965)]{forgy65}
Forgy, E.
\newblock Cluster analysis of multivariate data: Efficiency versus
  interpretability of classification.
\newblock \emph{Biometrics}, 21\penalty0 (3):\penalty0 768--769, 1965.

\bibitem[Frey \& Dueck(2007)Frey and Dueck]{affinity_propagation}
Frey, Brendan~J. and Dueck, Delbert.
\newblock Clustering by passing messages between data points.
\newblock \emph{Science}, 315:\penalty0 972--976, 2007.
\newblock URL \url{www.psi.toronto.edu/affinitypropagation}.

\bibitem[Hinton(2002)]{hinton2002training}
Hinton, Geoffrey~E.
\newblock Training products of experts by minimizing contrastive divergence.
\newblock \emph{Neural computation}, 14\penalty0 (8):\penalty0 1771--1800,
  2002.

\bibitem[LeCun et~al.(2015)LeCun, Bengio, and Hinton]{LeCun2015}
LeCun, Yann, Bengio, Yoshua, and Hinton, Geoffrey.
\newblock Deep learning.
\newblock \emph{Nature}, 521\penalty0 (7553):\penalty0 436--444, May 2015.
\newblock ISSN 0028-0836.
\newblock URL \url{http://dx.doi.org/10.1038/nature14539}.
\newblock Insight.

\bibitem[Likas et~al.(2003)Likas, Vlassis, and Verbeek]{likas2003global}
Likas, Aristidis, Vlassis, Nikos, and Verbeek, Jakob~J.
\newblock The global k-means clustering algorithm.
\newblock \emph{Pattern recognition}, 36\penalty0 (2):\penalty0 451--461, 2003.

\bibitem[MacQueen(1967)]{macqueen1967}
MacQueen, J.
\newblock Some methods for classification and analysis of multivariate
  observations, 1967.
\newblock URL \url{http://projecteuclid.org/euclid.bsmsp/1200512992}.

\bibitem[Rabiner(1989)]{Rabiner89atutorial}
Rabiner, Lawrence~R.
\newblock A tutorial on hidden markov models and selected applications in
  speech recognition.
\newblock In \emph{PROCEEDINGS OF THE IEEE}, pp.\  257--286, 1989.

\bibitem[Salakhutdinov \& Hinton(2009)Salakhutdinov and
  Hinton]{salakhutdinov2009deep}
Salakhutdinov, Ruslan and Hinton, Geoffrey~E.
\newblock Deep boltzmann machines.
\newblock In \emph{International Conference on Artificial Intelligence and
  Statistics}, pp.\  448--455, 2009.

\bibitem[Theodoridis \& Koutroumbas(2006)Theodoridis and Koutroumbas]{pam}
Theodoridis, Sergios and Koutroumbas, Konstantinos.
\newblock Chapter 14 - clustering algorithms iii: Schemes based on function
  optimization.
\newblock In Koutroumbas, Sergios~TheodoridisKonstantinos (ed.), \emph{Pattern
  Recognition (Third Edition)}, pp.\  589 -- 651. Academic Press, San Diego,
  third edition edition, 2006.
\newblock ISBN 978-0-12-369531-4.
\newblock \doi{http://dx.doi.org/10.1016/B978-012369531-4/50014-7}.
\newblock URL
  \url{http://www.sciencedirect.com/science/article/pii/B9780123695314500147}.

\bibitem[Vattani(2011)]{worst_kmeans}
Vattani, Andrea.
\newblock k-means requires exponentially many iterations even in the plane.
\newblock \emph{Discrete and Computational Geometry}, 45\penalty0 (4):\penalty0
  596--616, 2011.
\newblock ISSN 0179-5376.
\newblock \doi{10.1007/s00454-011-9340-1}.
\newblock URL \url{http://dx.doi.org/10.1007/s00454-011-9340-1}.

\bibitem[Zhang et~al.(2001)Zhang, Madigan, Moskewicz, and
  Malik]{zhang2001efficient}
Zhang, Lintao, Madigan, Conor~F, Moskewicz, Matthew~H, and Malik, Sharad.
\newblock Efficient conflict driven learning in a boolean satisfiability
  solver.
\newblock In \emph{Proceedings of the 2001 IEEE/ACM international conference on
  Computer-aided design}, pp.\  279--285. IEEE Press, 2001.

\bibitem[Zintchenko et~al.(2015)Zintchenko, Hastings, and Troyer]{groups}
Zintchenko, Ilia, Hastings, Matthew~B., and Troyer, Matthias.
\newblock From local to global ground states in ising spin glasses.
\newblock \emph{Phys. Rev. B}, 91:\penalty0 024201, Jan 2015.
\newblock \doi{10.1103/PhysRevB.91.024201}.
\newblock URL \url{http://link.aps.org/doi/10.1103/PhysRevB.91.024201}.

\end{thebibliography}

\end{document}